\documentclass{template/ecai}
\usepackage{times}
\usepackage{graphicx}
\usepackage{latexsym}
\usepackage{xcolor}
\usepackage{arydshln}
\usepackage{dblfloatfix}

\ecaisubmission   

\newcommand*\repeatinstitute[1][\value{footnote[1]}]{\footnotemark[#1]}

\begin{document}

\title{Refinement of Unsupervised \\ Cross-Lingual Word Embeddings}

\author{Magdalena Biesialska \institute{TALP Research Center, Universitat Polit\`ecnica de Catalunya, Barcelona, Spain, email: magdalena.biesialska@upc.edu} \and Marta R. Costa-juss\`a \repeatinstitute[1]}

\maketitle
\bibliographystyle{template/ecai}

\begin{abstract}
Cross-lingual word embeddings aim to bridge the gap between high-resource and low-resource languages by allowing to learn multilingual word representations even without using any direct bilingual signal. The lion's share of the methods are projection-based approaches that map pre-trained embeddings into a shared latent space. These methods are mostly based on the orthogonal transformation, which assumes language vector spaces to be isomorphic. However, this criterion does not necessarily hold, especially for morphologically-rich languages. In this paper, we propose a self-supervised method to refine the alignment of unsupervised bilingual word embeddings. The proposed model moves vectors of words and their corresponding translations closer to each other as well as enforces length- and center-invariance, thus allowing to better align cross-lingual embeddings. The experimental results demonstrate the effectiveness of our approach, as in most cases it outperforms state-of-the-art methods in a bilingual lexicon induction task.
\end{abstract}

\section{INTRODUCTION}
There are roughly 7000 languages around the world \cite{maxwell-hughes-2006}, and thus multilingual Natural Language Processing (NLP) has been a long-standing goal. Yet, NLP systems nowadays mainly support only the English language. This stems from a limited number of available parallel corpora even for resource-rich languages. The necessity to rely on bilingual data poses a great constraint on the development of multilingual NLP systems.

Cross-lingual word embeddings (CLEs) aim to bridge the gap between high-resource and low-resource languages by enabling to learn multi-lingual word representations even without any parallel data. More specifically, CLEs are representations of words in different languages, trained independently on monolingual corpora, and subsequently mapped into a shared vector space via linear transformation. Not surprisingly, CLEs have been attracting a lot of attention lately, as they allow to compare a word's meaning between languages and enable cross-lingual transfer learning \cite{ruder2017survey}. These properties are beneficial for resource-rich languages, but are even more desirable in low-resource scenarios. Hence, this makes CLEs useful in downstream NLP tasks, such as: bilingual lexicon induction, neural machine translation, document classification, and information retrieval, among others.

In fact, there exist various methods to obtain CLEs, where a key differentiator is the nature and amount of a bilingual signal provided during training. 

\paragraph{Supervised.} Early methods \cite{mikolov2013exploiting,faruqui-dyer-2014-improving} leveraged large, prepared in advance, bilingual dictionaries to learn cross-lingual embedding mappings. Later it was shown that the number of seed word translations can be reduced considerably \cite{vulic-korhonen-2016-role}, however, the requirement of bilingual supervision has remained the same.

\paragraph{Weakly supervised.} These bootstrapping approaches rely on typically small seed lexicons. In particular, CLE models that exploit a weak supervision, use initial bilingual seeds based on: cognates \cite{smith2017offline}, identical words \cite{sogaard2018limitations} or shared numerals \cite{artetxe2017acl}.

\paragraph{Unsupervised.} The most recent line of research  \cite{zhang-etal-2017-adversarial,artetxe2018acl,conneau2018word}, allows to learn CLEs without the need of any bilingual signal. Interestingly, CLEs trained solely on monolingual corpora are reported to demonstrate performance on a par with or even outperform supervised methods \cite{artetxe2018acl,conneau2018word,alvarez-melis-jaakkola-2018-gromov}.    
Importantly, Grave \textit{et al.} \cite{grave2018unsupervised} observe that refinement methods significantly improve the quality of weakly supervised and unsupervised CLE models. \\

In this work, we make a number of contributions. Firstly, we introduce a method for a self-supervised refinement of unsupervised CLEs. In contrast with existing approaches, our method is fully unsupervised
and leverages a small self-learned seed lexicon. In addition, to the best of our knowledge, we are the first to apply a self-supervised refinement method to the state-of-the-art\footnote{According to the results of a comparative study of cross-lingual embedding models presented in \cite{glavas-etal-2019-properly}.} unsupervised CLE model proposed by Artetxe \textit{et al.} \cite{artetxe2018acl}. Secondly, in this work, we address the problem of imperfect isomorphism in embedding vector spaces. Lastly, through the evaluation of our approach on a standard bilingual dictionary induction benchmark, we show that our method improves the word translation accuracy for almost all investigated language pairs. 

\section{RELATED WORK}
\label{sec:related_work}
\subsection{Isomorphic Vector Spaces and Orthogonal Transformations}
\label{sec:isomorphic_orthogonal}

 According to the popular claim stated by Mikolov \textit{et al.} \cite{mikolov2013exploiting}, an alignment between word vector spaces representing two different languages is possible, because same concepts in different languages bear similar statistical properties, and thus vector spaces of these languages can be considered isomorphic. Two graph spaces, such as words embeddings, are isomorphic if they contain the same number of graph (words) vertices (or for a relaxed version, only for the most frequent \textit{k} words) connected in the same way. Under this assumption, there have been many works \cite{artetxe2017acl,conneau2018word,ruder-etal-2018-discriminative} successfully utilizing orthogonal mapping methods to extract bilingual lexicons using CLEs. Importantly, such orthogonal transformations preserve length and inner products of vector representations of words.

The effectiveness of these orthogonal transformations falls drastically when the isomorphic condition does not hold. This problem can be observed for morphologically-rich or distant languages such as English and Japanese \cite{sogaard2018limitations,fujinuma-etal-2019-resource}, which in case of CLEs may be considered non-isomorphic language pairs \cite{zhang-etal-2019-girls}. Therefore, in this work, we report results for morphologically-rich languages such as German and Finnish (see Table \ref{tab:bdi-results}). 

\subsection{Cross-Lingual Mapping Methods}
\label{sec:mapping_methods}
This study only focuses on the methods based on word-level alignment. The vast majority of CLE models can be classified as \textbf{projection-based methods} (also referred as mapping-based) \cite{mikolov2013exploiting}. In this approach, word embeddings in two languages are trained independently on monolingual corpora, and next these word representations are mapped to a shared space using a linear transformation. 
The transformation matrix is usually learned from parallel data, such as word alignments or bilingual dictionaries \cite{ruder2017survey}. Clearly, this method is suitable for supervised and weakly-supervised settings. Nevertheless, recently unsupervised models have made a successful breakthrough, showing that monolingual corpora alone are sufficient for learning the transformation.

Other CLE models fall into two categories: \textbf{pseudo-mixing methods} and \textbf{joint approaches} \cite{ruder2017survey}. While the latter category of methods jointly optimizes monolingual and cross-lingual objectives, the former group does not rely on finding the mapping between the source and target language. Concretely, pseudo-mixing methods use word-level alignment from a seed dictionary to build a pseudo-bilingual corpus with source words being randomly replaced with their translations. 

Since our proposed model leverages the projection-based approach, we will not further discuss the other two aforementioned methods, but rather we will concentrate on the mapping-based approaches. Existing projection-based methods can be classified into four groups:
\begin{itemize}
    \item \textbf{regression methods} map the source language embeddings to the target language space using a least-squares objective \cite{mikolov2013exploiting,dinu2015improving,shigeto2015};
    \item \textbf{canonical methods} map the word representations in both languages to a new shared space using canonical correlation analysis \cite{faruqui-dyer-2014-improving,lu-etal-2015-deep,Ammar2016MassivelyMW};
    \item \textbf{orthogonal methods} map the source language embeddings to maximize the similarity with the target language representations under the constraint of orthogonal transformation \cite{xing-etal-2015-normalized,artetxe2016emnlp,smith2017offline,zhang-etal-2016-ten};
    \item \textbf{margin methods} map the source language embeddings to maximize the margin between the correct translations and other candidates \cite{lazaridou-etal-2015-hubness}.
\end{itemize}

 In fact, \cite{artetxe2018acl} demonstrated that regression, canonical and orthogonal methods can constitute a multi-step linear transformation framework.

\section{METHODOLOGY}
\begin{table*}
\begin{center}
{\caption{Bilingual Lexicon Induction results. Precision at k=1 (P@k x 100\%) performance for Spanish (ES), German (DE) and Finnish (FI), where English (EN) is a source language.}\label{tab:bdi-results}}
\begin{tabular}{lccccccccccccccc}
\hline
\rule{0pt}{12pt}
&\multicolumn{1}{l}{}&\multicolumn{2}{c}{\textbf{EN-ES}}&\multicolumn{1}{l}{}&\multicolumn{2}{c}{\textbf{EN-DE}}&\multicolumn{1}{l}{}&\multicolumn{2}{c}{\textbf{EN-FI}}\\
\cline{3-4}\cline{6-7}\cline{9-10}
\rule{0pt}{12pt}
& & P@1 & $\Delta$ & & P@1 & $\Delta$ & & P@1 & $\Delta$\\
\hline
\\[-6pt]
\multicolumn{2}{l}{\textsc{VecMap}} & 37.47 & & & 48.47 & & & 33.08\\
\multicolumn{2}{l}{unsup. \textsc{IterNorm + VecMap}} & 36.33 & -1.14 & & 48.47 & 0.00 & & 32.79 & -0.29 \\[+2pt]
\hdashline
\\[-7pt]
\multicolumn{2}{l}{Our method} & \textbf{37.67} & \textbf{+0.20} & & 48.47 & 0.00 & & \textbf{33.29} & \textbf{+0.21}
\\[+2pt]
\hline
\end{tabular}
\end{center}
\end{table*}

Our approach is motivated by the success of unsupervised CLE models, as well as recent promising results demonstrated by the refinement methods applied to supervised CLE models \cite{doval2018improving,zhang-etal-2019-girls}. In this work, we build upon these models; however, in contrast to the existing refinement approaches, we have designed our method to perform well in a more challenging unsupervised scenario.

The proposed method is based on the state-of-the-art approach introduced in \cite{artetxe2018acl}, and extends it by applying additional transformations to refine the CLEs. In that respect, we follow the idea of \cite{doval2018improving} to create a cross-lingual vector space, which corresponds to the average of the aligned source and target language spaces. However, their method is supervised, and thus uses bilingual lexicon when performing a refinement of the initial alignment. Our method, on the other hand, is fully unsupervised and leverages self-learned seed dictionary to map the source and target language embeddings onto their average. More concretely, the training process is composed of three steps.

Firstly, having monolingual corpora for both source and target languages, word representations are learned for each language independently. In this step, word embedding methods such as Word2Vec \cite{mikolov2013efficient}, GloVe \cite{pennington2014glove} or fastText \cite{bojanowski-etal-2017-enriching} can be applied to obtain the monolingual embeddings.

Secondly, the source and target language embeddings are mapped to a shared vector space by means of a linear transformation. However, following \cite{artetxe2018acl}, the vectors are normalized before the transformation is performed. This step normalizes the length of word vectors and performs dimension-wise mean centering. After this pre-processing step, embeddings can be aligned. While there exists a number of mapping methods (as described in Section \ref{sec:mapping_methods}), we will only explain approaches used in our model.
At the outset, an initial seed lexicon needs to be learned in a fully unsupervised way. Artetxe \textit{et al. }\cite{artetxe2018acl} employ a heuristic initialization method grounded in the idea that words in different languages have similar distributions, assuming that the embedding spaces are perfectly isomorphic (this is a simplification and in our proposed model we aim to fix it). Afterwards, the initial seed lexicon is improved using refinement methods. 

Finally, in the proposed model, the last step is a self-supervised refinement of the alignment that is applied after the initial mapping is done. In general, the proposed method is motivated by the assumption that vector spaces of source and target language embeddings have different structure and should not be considered entirely isomorphic. Hence, when we operate in a shared cross-lingual space it is evident that source embeddings and their translations are still distant. Therefore, our refinement method consists of two phases.

\paragraph{Averaging the vectors.} The underlying idea behind this step is to bring closer source words and their translations. Hence, for each word that is included in the induced dictionary, following the approach of \cite{doval2018improving}, we shift each embedding vector to reach the middle point between the source word and its translation. More specifically, the vector average is computed in a standard way:
\begin{equation}
\vec{\mu}_{w, w^{\prime}}=\frac{\vec{v}_{w}+\vec{v}_{w^{\prime}}}{2}
\end{equation}
where, $w \in V$ and $w^{\prime} \in V^{\prime}$ are source and target language words, and then the value is assigned to each embedding. To this end, we do not use any supervised source of parallel data, as the bilingual dictionary ${D = \{(w,w^{\prime})\}}$ is induced during the initial alignment step. 

\paragraph{Length normalization and mean centering.} As the entire projection-based unsupervised CLE method relies on the orthogonal assumption; therefore, we concur with \cite{zhang-etal-2019-girls} that word embeddings should be of the same unit length. Moreover, they stress the importance of source and target language vector spaces having equal magnitude centers. Therefore, every source and target word vector is transformed iteratively to fulfil both conditions:
\begin{equation}
\mathbf{y}_{i}^{(k)}=\mathbf{x}_{i}^{(k-1)} /\left\|\mathbf{x}_{i}^{(k-1)}\right\|_{2}
\end{equation}
and
\begin{equation}
\mathbf{x}_{i}^{(k)}=\mathbf{y}_{i}^{(k)}-\frac{1}{n} \sum_{i=1}^{n} \mathbf{y}_{i}^{(k)}
\end{equation}
respectively, where ${{x}_{i} \in \{(\vec{v}_{w},\vec{v}_{w^{\prime}})\}}$, ${\|{x}_{i}\|_{2}=1}$ for all $i$, and $\sum_{i=1}^{n}{x}_{i}={0}$.

\section{EXPERIMENTS}
\label{sec:experiments}
In this section, we evaluate the quality of the CLE models on a standard task of bilingual lexicon induction.

\subsection{Experimental Setup}

\paragraph{Data.}
We conduct our experiments using a popular dataset introduced by Dinu \textit{et al.} \cite{dinu2015improving} and its extensions \cite{artetxe2017acl,artetxe2018aaai}. The used dataset consists of the following language pairs: English-Spanish, English-German and English-Finnish. Monolingual embeddings of 300 dimensions were created using Word2Vec\footnote{https://code.google.com/archive/p/word2vec/} \cite{mikolov2013efficient} and were trained on WMT News Crawl (Spanish), WacKy crawling corpora (English, German), and Common Crawl (Finnish). To evaluate the performance, we use bilingual dictionaries provided in \textsc{VecMap}\footnote{https://github.com/artetxem/vecmap}, where each test set consists of 1500 entries.

\paragraph{Baselines.}
We report our results in comparison with \textsc{VecMap} in the unsupervised mode \cite{artetxe2018acl}. Furthermore, we compare our results with a recent refinement method \textsc{IterNorm}\footnote{We would like to thank the authors for sharing with us a code snippet with an implementation of their method.} \cite{zhang-etal-2019-girls} which, contrary to the original paper, is used here in the unsupervised setting. We perform the evaluation using \textsc{VecMap} scripts with CSLS method used for retrieval (instead of nearest neighbor).

\subsection{Bilingual Lexicon Induction}
The intrinsic task of bilingual lexicon induction is a common choice to evaluate CLE models. The goal of this task is to indicate the most appropriate translation for each source word, given nearest neighbor target embeddings in the shared vector space. The accuracy is measured as the percentage of correctly translated source words with respect to a ground truth translation from a dictionary.
This task is considered a good proxy for evaluating the performance of CLEs, as high-quality bilingual lexicons are available for many language pairs. However, one may argue that existing dictionaries merely contain the most frequent words and bilingual lexicon induction task should be accompanied by other evaluation methods \cite{glavas-etal-2019-properly} .

We follow the standard evaluation procedure by measuring scores for Precision at 1 (P@1), which determines how many times one of the correct translations of a source word is retrieved as the nearest neighbor of the source word in the target language. We report our results in Table \ref{tab:bdi-results}.


We observe that our method outperforms baseline models in two cases: English-Spanish and English-Finnish. Furthermore, it performs on a par with baselines for the English-German language pair. As it can be seen, our method obtains better scores than \textsc{IterNorm} in all cases but one. While the method proposed in \cite{zhang-etal-2019-girls} was originally trained in the supervised setting, it is universal and can be applied in the case of an unsupervised CLE model as well. Although it achieved very good results in the supervised setting, according to our experiments, it does not perform as good when combined with the unsupervised \textsc{VecMap} model. We hypothesize, that the reason why our method surpasses the baselines is mainly due to the use of self-learned dictionary, which improves subsequent transformations.


\section{CONCLUSION AND FUTURE WORK}
This work adds to the growing body of research in CLEs. First, we introduced a self-supervised method to refine unsupervised bilingual word embeddings by leveraging a small self-learned seed lexicon. To our knowledge, this was the first attempt to apply a self-supervised refinement method to the state-of-the-art unsupervised CLE model by Artetxe \textit{et al.} \cite{artetxe2018acl}. Second, our work addressed the problem of imperfect isomorphism in embedding vector spaces. The results, achieved in a bilingual dictionary induction task, suggest that our proposed approach improved the state-of-the-art for almost all evaluated language pairs. 

In the future we plan to investigate if our method boosts the performance of existing models in downstream tasks, especially in unsupervised neural machine translation. Moreover, it would be also interesting to experiment with adapting our refinement technique to a multilingual alignment setting to improve cross-lingual transfer. In addition, as traditional (context-invariant) word embeddings suffer from the meaning conflation deficiency, a study of cross-lingual embeddings in relation to unsupervised sense representations and contextual embeddings would be interesting to perform.

\ack 
We thank anonymous reviewers for their helpful comments.
This work is supported in part by the Spanish Ministerio de Econom\'ia y Competitividad, the European Regional Development Fund through the  postdoctoral  senior grant Ram\'on y Cajal and by the and the Agencia  Estatal  de  Investigaci\'on through the projects EUR2019-103819 and PCIN-2017-079.

\bibliography{REFERENCES}

\end{document}